%% file: main.tex
\documentclass[sigconf]{acmart}
\AtBeginDocument{%
  \providecommand\BibTeX{{%
    \normalfont B\kern-0.5em{\scshape i\kern-0.25em b}\kern-0.8em\TeX}}}

\copyrightyear{2022} 
\acmYear{2022} 
\setcopyright{acmcopyright}\acmConference[BCB '22]{13th ACM International Conference on Bioinformatics, Computational Biology and Health Informatics}{August 7--10, 2022}{Northbrook, IL, USA}
\acmBooktitle{13th ACM International Conference on Bioinformatics, Computational Biology and Health Informatics (BCB '22), August 7--10, 2022, Northbrook, IL, USA}
\acmPrice{15.00}
\acmDOI{10.1145/3535508.3545518}
\acmISBN{978-1-4503-9386-7/22/08}

\usepackage[figuresright]{rotating}
\usepackage{subfigure}
\usepackage{hyperref}       
\usepackage{url}            
\usepackage{booktabs}       

\usepackage{graphicx}
\usepackage{algpseudocode}
\usepackage{xspace}
\usepackage{tabu}
\usepackage{bm}
\usepackage{multirow}
\usepackage{enumitem}
\usepackage[ruled,vlined,linesnumbered]{algorithm2e}
\usepackage{color,soul}
\usepackage{wrapfig}
\usepackage{amsmath,multicol, mathrsfs}
\usepackage[bottom]{footmisc}
\usepackage{makecell}

\begin{document}


\title{Attention-based Aspect Reasoning for Knowledge Base Question Answering on Clinical Notes}

\author{Ping Wang$^1$, Tian Shi$^2$, Khushbu Agarwal$^3$, Sutanay Choudhury$^3$, Chandan K. Reddy$^2$}
\affiliation{$^1$Dept of Computer Science, Stevens Institute of Technology, Hoboken, NJ\country{}}
\affiliation{$^2$Dept of Computer Science, Virginia Tech, Arlington, VA $^3$Pacific Northwest National Laboratory, Richland, WA\country{}}
\email{ping.wang@stevens.edu,researchtianshi@gmail.com,{khushbu.agarwal, sutanay.choudhury}@pnnl.gov, reddy@cs.vt.edu}

\renewcommand{\shortauthors}{P. Wang, et al.}

\begin{abstract}
Question Answering (QA) in clinical notes has gained a lot of attention in the past few years. Existing machine reading comprehension approaches in clinical domain can only handle questions about a single block of clinical texts and fail to retrieve information about multiple patients and their clinical notes. To handle more complex questions, we aim at creating knowledge base from clinical notes to link different patients and clinical notes, and performing knowledge base question answering (KBQA). Based on the expert annotations available in the n2c2 dataset, we first created the ClinicalKBQA dataset that includes around 9K QA pairs and covers questions about seven medical topics using more than 300 question templates. Then, we investigated an attention-based aspect reasoning (AAR) method for KBQA and analyzed the impact of different aspects of answers (e.g., entity, type, path, and context) for prediction. The AAR method achieves better performance due to the well-designed encoder and attention mechanism. From our experiments, we find that both aspects, type and path, enable the model to identify answers satisfying the general conditions and produce lower precision and higher recall. On the other hand, the aspects, entity and context, limit the answers by node-specific information and lead to higher precision and lower recall.
\end{abstract}

\ccsdesc[500]{Mathematics of computing~Graph algorithms}
\ccsdesc[500]{Computing methodologies~Neural networks}
\ccsdesc[500]{Computing methodologies~Learning latent representations}
\ccsdesc[500]{Computing methodologies~Information extraction}

\keywords{Clinical knowledge base, question answering, aspect representation, attention mechanism.}


\maketitle
\input{1-introduction}

\input{2-related}

\input{3-model}
\input{4-experiments}
\input{5-conclusion}

\begin{acks}
This work was supported in part by the US National Science Foundation grant IIS-1838730, Amazon AWS cloud computing credits, and Pacific Northwest National Laboratory under DOE-VA-21831018920.
\end{acks}

\bibliographystyle{ACM-Reference-Format}
\bibliography{ref}

\newpage
\appendix
\input{6-appendix}

\end{document}

%% file: 1-introduction.tex
\section{Introduction} 
\label{kbqa:sec1}
Electronic Health Records (EHR) provide comprehensive information that can assist doctors with their clinical decision making.
Traditionally, doctors retrieve the information of patients via accessing structured databases with rule-based systems and reading their clinical notes.
Recently, several attempts have been made to build Question-Answering (QA) systems on EHR \cite{pampari2018emrqa,vsuster2018clicr,wang2020text} so that doctors can get answers for their questions more efficiently.
Generally speaking, QA systems can be grouped into several categories according to the format of data sources. 
For example, machine reading comprehension (MRC) performs QA on plain text data \cite{rajpurkar2016squad}. 
Text-to-SQL problem performs QA on database \cite{yu2018spider,zhong2017seq2sql}.
Knowledge Base QA (KBQA) \cite{bordes2014open} aims at finding answers from the underlying Knowledge Base (KB), such as Freebase~\cite{bollacker2008freebase}.
In our previous work \cite{wang2020text}, we introduced a MIMICSQL dataset for Text-to-SQL generation on MIMIC III database \cite{wang2020text}, which is limited to retrieving answers from a database, which does not cover information that are not quantified or structured, such as family history and discharge conditions.
Pampari\textit{ et al.} \cite{pampari2018emrqa} proposed an emrQA dataset for MRC on clinical notes.
However, it only supports the task of accessing information from a single block of text, which is not practical for doctors who may be interested in retrieving information from a collection of clinical notes.

In this work, we present ClinicalKBQA, a dataset for QA on clinical KB (ClinicalKB) constructed from clinical notes, which alleviates the problems encountered with emrQA by allowing doctors to access information across different notes.
ClinicalKBQA is composed of two subsets, namely, Clinical Knowledge Base (ClinicalKB) and Question-Answering (QA) pairs, both of which are constructed by leveraging existing annotations of clinical notes that are available for various NLP tasks in n2c2\footnote{https://n2c2.dbmi.hms.harvard.edu/} (previously known as i2b2).
ClinicalKB integrates advantages of both structured database and unstructured clinical notes. 
On the one hand, the intrinsic graph structure of ClinicalKB connects the information of different patients and clinical notes via relations/edges, which allows it to answer questions associated with many patients and clinical notes (e.g., \textbf{Q3, Q4} in Table~\ref{tab:question-example}). 
On the other hand, ClinicalKB includes comprehensive patient information as in clinical notes, which makes it possible to answer questions not covered in database (e.g., \textbf{Q2, Q4} in Table~\ref{tab:question-example}).

\begin{table}[!tp]
\centering
\caption{Comparisons of the answerable questions over different types of EHR, including Clinical Notes (CN), Structured Tables (ST) and Knowledge Base (KB). The symbol ``\checkmark" indicates that the questions are answerable.}
\resizebox{\linewidth}{!}{
\begin{tabular}{p{6.5cm}|p{0.19cm}p{0.19cm}p{0.19cm}}\hline 
\bf Questions & \bf CN & \bf ST & \bf KB\\ \hline

\textbf{Q1}: What medications has patient P939003 ever been prescribed? &\checkmark &\checkmark &\checkmark \\
\hline

\textbf{Q2}: What does patient P961115 take ibuprofen for? &\checkmark & &\checkmark \\
\hline

\textbf{Q3}: Which patients have been diagnosed with both Gout and GERD? & &\checkmark &\checkmark \\
\hline

\textbf{Q4}: What are the obese indicators of heart disease in all medical records of patient P258? & & &\checkmark \\
\hline
\end{tabular}
}
\label{tab:question-example}
\vspace{-3mm}
\end{table}

To tackle the KBQA challenges in ClinicalKBQA dataset, we investigated an attention-based aspect reasoning (AAR) approach. Specifically, for each input question, we represent each candidate answer as four aspects, including entity, type, path, and context, and analyze the matching scores between the input question and candidate answers based on their embeddings. 
Through the analysis of our results, we found that the impact of different candidate aspects on retrieving final answers tends to be different. Two aspects, entity and context, provide the node specific information, which helps to retrieve nodes that satisfy the constraints specified in the questions. While the general information included in the other two aspects, type and path, is helpful for the model to filter out more nodes that satisfy the constraints about the node type and path.
In summary, our \textit{major contributions} are: (1) Created a dataset for knowledge base question answering in healthcare. It consists of two sets: (i) \textit{ClinicalKB}: a comprehensive clinical knowledge base created based on the expert annotations in n2c2 dataset, and (ii) \textit{QA pairs}: a large-scale question answering dataset on ClinicalKB. 
(2) Investigated an attention-based aspect-level reasoning (AAR) method for KBQA. 
(3) Conducted experimental analysis on ClinicalKBQA dataset to analyze the performance of AAR model and the significance of different aspects in providing accurate answers. 




%% file: 2-related.tex
\section{Related Works}
\label{kbqa:sec2}

Question-Answering (QA) aims at automatically answering natural language questions about data sources in a variety of formats, including free text~\cite{rajpurkar2016squad}, knowledge base~\cite{cui2019kbqa}, and database~\cite{zhong2017seq2sql}.
Knowledge base question answering (KBQA) has gained a lot of attention in recent years with the rapid growth of large-scale knowledge bases, such as YAGO2~\cite{hoffart2011yago2} and Freebase~\cite{bollacker2008freebase}.
Advances in deep neural networks also allowed KBQA models to be trained in an end-to-end manner \cite{bordes2014question,hao2017end,chen2019bidirectional} and achieve competitive performance compared to traditional semantic parsing based methods \cite{abujabal2017automated,kwiatkowski2013scaling}.

QA in the healthcare domain is still an underexplored research topic, especially due to the lack of large-scale annotated datasets and patient privacy issues \cite{jin2021biomedical}. 
Traditional biomedical QA depends on rule-based or heuristic feature-based methods~\cite{athenikos2010biomedical}.
Recently, several datasets have been created for machine reading comprehension (MRC), including BioASQ for semantic indexing and QA~\cite{tsatsaronis2015overview}, CliCR for MRC on clinical case reports~\cite{vsuster2018clicr}, PubMedQA for MRC on biomedical research texts~\cite{jin2019pubmedqa} and emrQA for MRC on clinical notes~\cite{pampari2018emrqa}. 
MIMICSQL~\cite{wang2020text} was created for QA on structured EMR data by translating questions to SQL queries. 
These datasets allow researchers to handle unique challenges present in the healthcare domain.
There are several works about knowledge base in healthcare.
SNOMED~\cite{donnelly2006snomed} is a KB with standard clinical terminologies for healthcare documentation. Unified Medical Language System (UMLS) \cite{bodenreider2004unified} is an integration of medical terminology, classification and coding standards including SMOMED.
Rotmensch \textit{ et al.} \cite{rotmensch2017learning} learnt a knowledge graph of symptom and disease from EMR by considering the importance measure between terms.

For KBQA modeling, Generally speaking, there are two groups of methods \cite{lan2021survey}, i.e., semantic parsing-based \cite{liang2017neural,berant2014semantic,reddy2014large}  and information retrieval-based (IR-based) \cite{yao2014information,bordes2015large,dong2015question,chen2019uhop} methods. 
Semantic parsing-based methods parse the input questions into a logical format, which is the syntactic representation of the input questions. To predict the answers, the logical format is aligned with the KB structures and further executed against the KB.
IR-based approaches first directly identify and rank the candidate answers from the KB by considering the information in the natural language questions, and then perform the reasoning by learning the representation of the input questions and analyzing the semantic matching of the questions and candidate answers.
As an important category of IR-based KBQA methods, embedding-based approaches \cite{bordes2014question,hao2017end} map questions and answer candidates onto a common embedding space and directly calculate their matching scores. Then, ranking techniques are adopted to search answers from KB for given questions. The survey papers \cite{diefenbach2018core,lan2021survey} provide a comprehensive analysis and summary about the KBQA task.

%% file: 3-model.tex
\section{The ClinicalKBQA Dataset}
\label{kbqa:sec3}

ClinicalKBQA consists of two subsets, i.e., ClinicalKB and QA pairs. In this section, we will explain how we created the clinical knowledge base and the question answering dataset.

\subsection{ClinicalKB}
The n2c2 challenge data provide fine-grained document-level expert annotations of clinical records for various NLP tasks in clinical domain. We leverage the annotations about seven tasks to build the clinical knowledge base,
including smoking status classification \cite{uzuner2008identifying}, identification of obesity and its comorbidities \cite{uzuner2009recognizing}, medication extraction \cite{uzuner2010extracting}, relations extraction~\cite{uzuner2010community}, co-reference resolution \cite{uzuner2012evaluating}, temporal information extraction~\cite{sun2013evaluating}, and risk factors prediction~\cite{stubbs2015identifying}.
The narrative blocks in clinical notes, such as family history, provide more detailed clinical information from different aspects and can be efficiently extracted with rule-based methods as additional annotations.

Grounded on domain expert annotated clinical notes in the n2c2 challenge data, we construct clinical KB following two steps: (1) \textit{Identify entities.}
An entity is represented by its name and type. 
(2) \textit{Build triples}, i.e., (subject, predicate, object). Here, both subject and object are entities, and predicate is a relation between them.
In addition, we have also fixed some problems in the annotations during pre-processing, such as pronouns like ``this/that/his/her" and irrelevant punctuation.

\begin{table*}[!t]
\centering
\caption{Statistics of ClinicalKB and QA pairs created based on the n2c2 dataset. Here, QuesLen, GoldAns, and CandAns represent question length, gold-standard answers, and candidate answers, respectively.}
\vspace{-1.5mm}
\resizebox{1.0\linewidth}{!}{
\begin{tabular}{l|ccccccc}\hline 
\bf Metric & \bf Smoking & \bf Obesity & \bf Medications &\bf Relations &\bf Co-reference &\bf Temporal &\bf Risk \\ \hline

 \# Patients &502 &1,103 &261 &426 &424 &310 &119 \\
\# Entities &6,160 &17,861 &28,821  &20,031 &1,581 &127,772 & 6,984\\
\# Entity types &49 &42 &46  &7 &7 &20 &15 \\
\# Triples &9,730 &42,474 &53,519  &30,401 &1,378 &276,513 &24,553 \\
\# Relations &5 &8 &14  &11 &7 &13 &11 \\
\hline
\# Question Templates &26 &37 &59  &74 &18 &29 &79 \\
\# QA pairs &600 &1,126 &1,847  &2,389 &444 &626 &1,920 \\
Min/Max/Avg QuesLen &4/10/8 &5/14/9 &5/17/10  &6/21/11 &8/17/12 &8/19/11 &8/21/17 \\
Min/Max/Avg \# GoldAns &1/82/5 &1/816/27 &1/111/10  &1/29/3 &1/2/2 &1/239/19 &1/69/5\\
Min/Max/Avg \# CandAns &5/2,665/999 &3/8,686/2,261 &2/6,240/68  &4/679/79 &3/6/4 &5/1,543/175 &2/74/17 \\
\hline
\end{tabular}
}
\label{tab:data-stat}
\vspace{-2.5mm}
\end{table*}

\subsection{Question-Answer (QA) Pairs}

\subsubsection{Question Collection}
{We first collect a set of questions by polling real interests of physicians and considering existing clinical question resources, including emrQA and MIMICSQL, and further identify questions that can be answered by ClinicalKB.}
Compared with QA on structured tables \cite{wang2020text} and clinical notes \cite{pampari2018emrqa}, the questions on ClinicalKB cover a much wider range of topics~(see Table~\ref{tab:question-example}).
Some questions are not answerable by structured tables or a single clinical note.
Take \textbf{Q4} as an example, ``indicators of diseases'' are usually not included in structured tables, and the term ``all medical records'' indicates that answers cannot be found in a single note. 

We then manually identified specific entities in the selected questions and replace them with generic placeholders to normalize and form question templates.
In total, we generated a set of $322$ question templates, including various paraphrases of questions with the same meanings. For example, the template for \textbf{Q2} in Table~\ref{tab:question-example} is ``\textit{What does patient $|$PatientID$|$ take $|$Medication$|$ for}?", where the generic placeholders \textit{$|$PatientID$|$} and \textit{$|$Medication$|$} are the topic entities of the question that need to be replaced by the corresponding ClinicalKB entities during question generation.
We believe that the questions we collected from domain experts and the existing clinical question sources recognized by the community will provide a helpful resource of QA for researchers in the scientific community.

\subsubsection{QA Pairs Generation.}
This step focuses on populating question templates and identifying corresponding answers. 
Since patient private information is de-identified in n2c2, we use patient IDs 
instead of names in patient-specific questions.
Each question template may have multiple ways to populate.
For example, the template of \textbf{Q2} mentioned previously can be populated with different combinations of \textit{$|$PatientID$|$} and \textit{$|$Medication$|$}. 
However, we do not need to enumerate all possible questions for it.
In practice, we applied two constraints to limit repetitions: 
(1)~Set a threshold to the total number of questions generated for each template.
(2)~Remove questions without answers. 
When generating questions, the corresponding answers to each question is simultaneously extracted from clinical notes based on the human annotation and ClinicalKB.

\subsection{Data Analysis}
\subsubsection{Basic Statistics.} 
The statistics of ClinicalKB and QA pairs are presented in Table~\ref{tab:data-stat}.
The ClinicalKB covers seven important medical topics in n2c2.
The total number of QA pairs is 8,952. 
We created more question templates and QA pairs for \textit{Medications}, \textit{Relations}, and \textit{Risk} because their annotations are more comprehensive.
The average question length is 12 in terms of tokens. 
Each question has at least one gold-standard answer and a lot of questions have multiple answers.
In this work, we refer to the collection of ClinicalKB and QA pairs as the ClinicalKBQA dataset. 
The number of entities in golden and candidate answers are 9 and 402 on average, respectively. The number of golden and candidate answers for questions about \textit{Co-reference} is relatively small since the variety of annotated terms with the same meaning are small in n2c2.

\subsubsection{Question Types. } Our primary goal of knowledge base question answering on clinical notes is to extract patient information from unstructured clinical text. Therefore, all questions included in our ClinicalKBQA dataset are factoid questions which aim to seek reliable and concise medical history information about patients. 
We analyzed the quantitative percentage of various question types in ClinicalKBQA data and find that the questions starting with ``What'', ``List/Search/Give/Provide'', and ``Which'' account for a large proportion of the dataset and aim to ask for detailed medical facts, such as prescribed medications and the smoking status. 
The questions starting with ``Why'' and ``How'' tend to be open-ended in many open-domain question answering datasets. However, in the ClinicalKBQA dataset, the ``Why'' and ``How'' types of questions are mainly included for retrieving attribute facts about medication, including prescribed reason, dosage, frequency, and duration. In addition, the question type ``When'' are included for extracting the admission and discharge time of patients.

\subsubsection{Question Coverage.}
Table~\ref{tab:question-example} provides a comparison of questions that can be answered on different types of EHR data. 
We can observe that knowledge base about patient clinical information is able to answer the basic questions that are answerable by QA on both clinical notes and structured tables. It also has the ability to combine the advantages of free-text clinical notes and structured tables to handle more complex questions. For example, 
for the question \textit{ Give me all diseases that are revealed by non contrast head ct scan on patient P0126}, even if there are lab test information included in the structured data, the diseases that are actually revealed by each test are not specified. While, for question \textbf{Q3} in Table~\ref{tab:question-example}, the machine reading comprehension on emrQA cannot provide answers since these questions are related to multiple clinical notes. ClinicalKB is able to integrate the information from different clinical notes or about different patients into a general network structure, which makes it feasible to handle more complex type of questions.

\section{The KBQA Modeling}
\label{kbqa:sec4}
\subsection{Candidate Generation}

It will be computationally expensive for KBQA models to directly search answers from ClinicalKB. Therefore, we first generate a candidate subgraph for each question in two steps:
(1)~We identify one of the entities in the question template as the topic entity (root), and collect all entities connected to it within 3-hop as a candidate subgraph.
Each entity in the subgraph except the root is viewed as a candidate answer.
For the ClinicalKBQA dataset, the answers to all questions are reachable within 3-hop of their topic entities.
(2)~We treat the remaining entities in the question as constraints to the candidate sub-graph, and further prune the graph to ensure that paths to the topic entity satisfy the constraints and include entities with expected answer type.
We show the basic statistics of candidate answers in Table~\ref{tab:data-stat}.

\vspace{-5mm}
\subsection{Attention-based Aspect Reasoning (AAR)}

Motivated by \cite{bordes2014question,hao2017end}, we implemented an embedding-based end-to-end model on ClinicalKBQA dataset that incorporates an attention mechanism between question representations and aspect-level answer candidate representations to calculate matching scores. 
There are mainly four components in the AAR model. 

\textbf{Question Representations.} 
The question encoder is composed of a word-embedding layer followed by a bi-directional LSTM layer, which encodes a question $q=(q_1,q_2,…,q_{|q|} )$ into a sequence of hidden states $H^q=(h_1^q,h_2^q,…,h_{|q|}^q )$, where $q_i$ and $h_i^q$ represent the $i^{th}$ token and its corresponding hidden state, respectively. 
$|q|$ is the length of the input question.    

\textbf{Graph Representation.} To encode candidate subgraphs, we first convert each subgraph to a candidate answer set, where each element (i.e., node) in the set represents an entity in the subgraph, which has four aspects of information: 1) \textit{Entity} ($h_k^e$) represents the embedding of the node $k$ in a KB. 2) \textit{Type} ($h_k^t$)  denotes the entity type of the node $k$. It provides important clue for finding an answer. 3)~\textit{Path} ($h_k^p$) represents a path from the topic entity node of a subgraph to the current candidate node. Thus, the path provides relationships between the topic entity and the candidate answer. 4) \textit{Context} consists of all neighboring nodes of the current candidate node $k$. We encoded the context of each candidate answer $c_k=(c_{k_1},c_{k_2}  ,…,c_{k_c})$ as a list of hidden states $H_k^c=(h_{k_1}^e,h_{k_2}^e,…,h_{k_c}^e )$, where $c_{k_i}$ is the $i^{th}$ context node in the context of node $k$, and  $h_{k_i}^e$ represents its corresponding entity embedding. For simplicity, we will use $h^e$, $h^t$, $h^p$, and $H^c$ to represent $h_k^e$, $h_k^t$, $h_k^p$, and $H_k^c$, respectively. 


\textbf{Attention Mechanisms.} 
The attention mechanisms  can discover underlying correlations between a question and different features/aspects of any candidate node. Here, we will use the attention between aspect ``type" and the input question, namely type-to-question attention, to illustrate how the attention mechanism works. Given the question representation $H^q=(h_1^q,h_2^q,…,h_{|q|}^q)$, and the type embedding $h^t$, the alignment score $u^{t2q}$ and attention weight $\alpha^{t2q}$ are calculated as $u_i^{t2q}=(h^t)^T tanh(W_{t2q}h_i^q+b_{t2q})$ and $\alpha_i^{t2q} =\frac{exp(u_i^{t2q})}{\sum_{j=1}^{|q|} exp(u_j^{t2q})}$, respectively.
Here $W_{t2q}$ and $b_{t2q}$ are model parameters. Finally, the type-related question representation, namely type-to-question representation, is obtained by $r^{t2q}=\sum_{j=1}^{|q|}\alpha_j^{t2q} h_j^q$
where $r^{t2q}$ is a question representation which incorporates type information. Similarly, we can obtain such representations for other aspects, including path, entity, and context. Hereafter, they are denoted as $r^{e2q}$, $r^{p2q}$ and $r^{c2q}$, respectively.

\textbf{Scoring Answers.} The prediction of answers is made based on the similarity score between the input question and each answer candidate, which is a weighted average score of distances between questions and different answer aspects of each candidate. For each aspect, we first calculate the similarity of its embedding and aspect-to-question representation as $s_{tq}=(h^t)^T r^{t2q}$. 
Since different aspects of candidate answers are not equally important to the final predictions, we also calculate the weight of each aspect as $w_{tq}=(H_{avg}^q)^T r^{t2q}$,
where $H_{avg}^q$ represents the question representation obtained by performing average-pooling over the sequence of hidden states of the question $H^q=(h_1^q,h_2^q,…,h_{|q|}^q )$. Therefore, the final score of each candidate answer will be $S(q,a)=w_{eq} s_{eq}+w_{tq} s_{tq}+w_{pq} s_{pq}+w_{cq} s_{cq}$.
During testing, candidate answers are ranked based on their scores.

\subsubsection{Training and Inference.} In the ClincalKBQA task, we treat the answer retrieval problem as a ranking problem and adopt a pair-wise strategy to train the model. Intuitively, ground truth answers should have higher scores than the other candidate answers. Therefore, during training, for each ground-truth answer node $a$ (positive example), we randomly select a candidate node (not an answer) $a'$ as a negative example. The training loss is a max-margin hinge loss defined as $L=min \frac{1}{|B^q|} \sum_{(a,a^{'})\in B^q} [\gamma+S(q,a)-S(q,a^{'})]_+$,
where $S(q,a)$ and $S(q,a')$ are the final scores of $a$ and $a'$, respectively. $\gamma \in (0,1)$ is a real number that indicates the margin between the positive and negative examples. $[\cdot]_+$ represents the hinge loss, which is defined by $max(0, \cdot)$.
Here, $B^q$ denotes a set of positive-negative example pairs $(a,a^{'})$, and $|B^q|$ is the batch size. Intuitively, the hinge loss function increases the margin between the positive and negative examples and allows us to select multiple answers from a set of candidate answers instead of the best answer only.

During the testing, for each input question, we first retrieve a set of candidate answers $C^q$ from the corresponding knowledge base, and then calculate the score for each candidate answer $a\in C^q$. The best answer is obtained by $a_{best}=arg\ \max_{a\in C^q} S(q,a)$.
Usually, there are multiple answers for each question, therefore, the candidate answers whose scores are close to the highest score within a margin can also be considered as answers. This inference process can be formulated as $f(q,a)=1$ if $S(q,a)>S(q,a_{best})-\gamma$. Otherwise,  $f(q,a)=0$.
Here $f(q,a)=1$ indicates that node $a$ is the answer to the question $q$.

%% file: 4-experiments.tex
\section{Experiments and Analysis}
\label{kbqa:sec5}

\subsection{Experimental Settings}
We implemented the AAR model along with its several variants. Following prior work, we adopted micro-averaged precision, recall, and F1 score to evaluate different models. 
In our experiment, we split the data into training/development/testing sets with a proportion of 5952/1000/2000. We implemented the AAR model using Pytorch \cite{paszke2017automatic} and the best set of parameters are selected based on the development set. We set the size of embeddings for words, entities, entity types, and paths to topic entities to 300. The word embeddings are learnt from scratch. We adopted a single layer Bi-LSTM with the hidden size 150. All parameters were trained using ADAM optimizer \cite{kingma2014adam} with a constant learning rate of 0.0001 for 10 epochs.
In addition, we compare the performance of AAR with a subgraph-based approach SGEmd \cite{bordes2014question}, which first calculates embeddings of words, entities, and path to topic entities. Then, each question representation is obtained by applying average pooling to word embeddings. Answer candidates are represented by entities, paths to topic entities, and subgraphs. This method is known as subgraph embedding. The ClinicalKBQA dataset and our implementation is made publicly available at this website\footnote{https://github.com/wangpinggl/Clinical-KBQA}.

\begin{table}[!tp]
    \centering
    \caption{Performance results on ClinicalKBQA using different evaluation metrics. Here, ``full'' indicates the model considering all the four aspects. While the rows with specific aspects indicate the model only considers these specific aspects.}
    \vspace{-1mm}
    \resizebox{1.05\linewidth}{!}{
    \begin{tabular}{l|ccccc}
        \hline    
        \bf Models & \bf Precision & \bf Recall & \bf Accuracy & \bf Micro-F1 & \bf Macro-F1  \\\hline
        \textbf{SGEmb} (full) & 0.7447 & 0.1249 &0.5105 & 0.2139 &0.6617
        \\
        \ \ \textit{entity $\&$ sub-graph} & 0.3866 & 0.0750 &0.2370 & 0.1256 &0.3807
        \\
        \ \ \textit{path} &  0.4870 & 0.7966 &0.7260 & 0.6045 &0.8583
        \\\hline
        \textbf{AAR} (full) & \bf 0.8072 & 0.1735 &0.6525 & 0.2856 &0.7973
        \\
        \ \ \textit{entity $\&$ context}  & 0.5964 & 0.1360 &0.4725 & 0.2216 &0.6645
        \\
        \ \ \textit{type} & 0.1205 &\bf 0.8908 &0.3685 & 0.2123 &0.6177
        \\
        \ \ \textit{path} & 0.4780 & 0.7913 &0.7665 & 0.5960 &\bf 0.9057
        \\
        \ \ \textit{type $\&$ path}  & 0.6598 & 0.6616 &\bf 0.7745 & \bf 0.6607 &0.8980
        \\\hline
    \end{tabular}
    }
    \label{tab:performance}
    \vspace{-3.5mm}
\end{table}

\begin{figure}[!tp]
\centering
\includegraphics[width=0.5\textwidth]{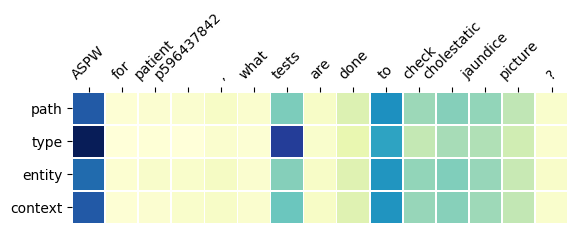}\label{fig:heat-map-2}
\vspace{-5mm}
\caption{Attention heatmaps generated by the cross-attention module. ASPW denotes weights for different aspects, which are \textit{path}, \textit{type}, \textit{entity}, and \textit{context}.}
\label{fig:heat-map}
\vspace{-2mm}
\end{figure}

\begin{figure}[!tp]
	\centering
	\includegraphics[width=1\linewidth]{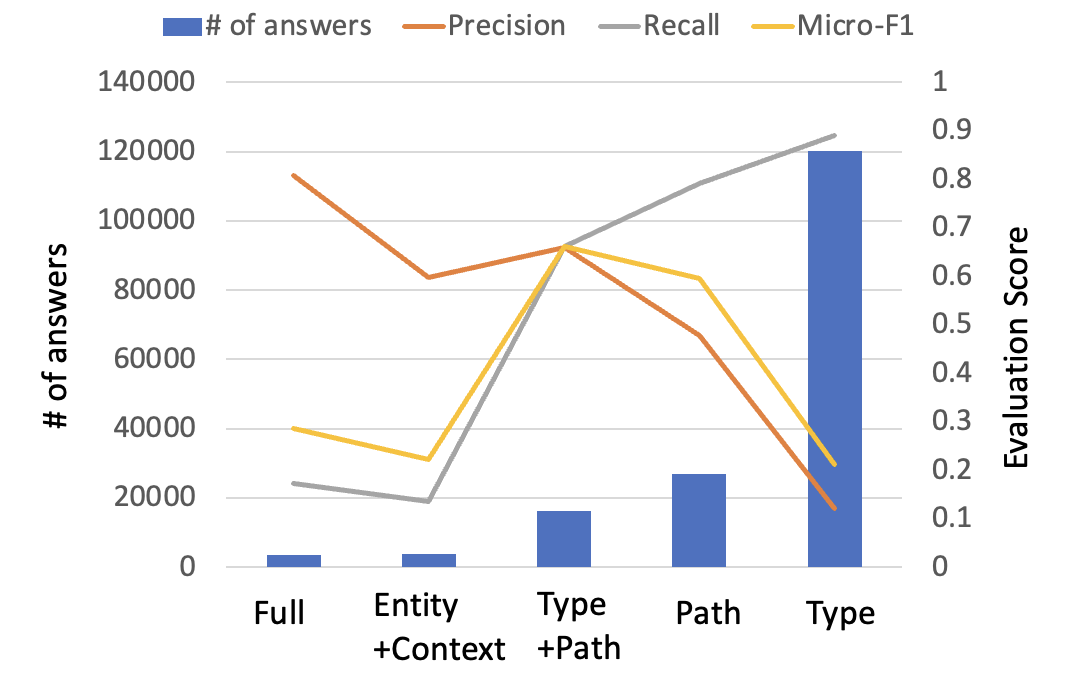}
	\caption{The impact of different aspects on the AAR model prediction, including the number of predicted answers and multiple evaluation scores.  \textbf{$\#$ of answers} denotes the number of answers predicted by models on testing set. The number of ground-truth answers is 16,251. The results show that the aspects \textit{type} and \textit{path} include general node information and lead to more answers that satisfy such generic requirements. However, the aspects \textit{entity} and \textit{context} include node specific information and lead to less number of answers that only satisfy such specific requirements.} 
	\label{fig:results-trend}
	\vspace{-3mm}
\end{figure}

\vspace{-2mm}
\subsection{Experimental Results}
From Table~\ref{tab:performance} and Figure \ref{fig:results-trend}, we can observe that AAR achieves better results than SGEmb, which is because AAR is equipped with a better encoder, attention mechanisms, and entity \textit{type} information.
To explore the impact of each aspect in our ClinicalKBQA, we studied models with only one aspect of information.
SGEmb-\textit{entity $\&$ sub-graph} and AAR-\textit{entity $\&$ context}, which only leverage entity embeddings, achieve relatively higher precision and lower recall, and the number of predicted answers is much fewer than that of the ground-truth.
This is due to the fact that different answer candidates have different entity embeddings and matching scores, which makes the model favor the answer with the highest score.

On the other hand, models that only consider \textit{path} and \textit{type} achieve relatively lower precision and higher recall since different candidates may share common \textit{type} and \textit{path} embeddings and have the same matching score.
Thus, the number of predicted answers are much more than that of the ground-truth.
Models with only \textit{path} information perform significantly better than other variants with only one aspect, which indicates \textit{path} is the most significant factor for our ClinicalKBQA.
Finally, AAR-\textit{type $\&$ path} achieves the best accuracy and Micro-F1 score.

We have also shown the heatmap based on the attention mechanism for an input question in Figure~\ref{fig:heat-map}. 
The model gives more weight to the aspect ``type" among all four aspects, which indicates that the aspect “type" of candidate answers is the most important feature for the final prediction. For aspect-towards-question attention, all four aspects capture the keywords ``tests" and ``to check cholestatic jaundice picture". These important keywords are serving as the query conditions to identify qualified candidate answers whose node ``type" is “test" and can be used “to check cholestatic jaundice picture". 
This analysis of attention weights is helpful for us to explain how the AAR model identifies correct answers for an input question. It also provides us insights about the impactful aspects of candidate answers to match the input questions on the ClinicalKBQA dataset.

%% file: 5-conclusion.tex
\vspace{-2mm}
\section{Conclusions}
\label{kbqa:sec6}

In this work, we introduced a dataset for question answering (QA) on ClinicalKB, namely ClinicalKBQA, which is composed of two subsets, i.e., ClinicalKB and QA pairs.
ClinicalKB is built from expert annotated clinical notes; thus, it allows doctors to ask questions on a collection of notes for different patients.
We have also introduced a procedure for generating answer candidate subgraphs from ClinicalKB for given questions.
In addition, an attention-based aspect-level reasoning model is investigated for KBQA on this newly created dataset. 
Finally, we conducted experimental analysis and studied the significance of different aspects in providing accurate answers. Based on the results, we find that KBQA can provide more accurate answers and cover more complex questions.
In addition, the aspects \textit{type} and \textit{path} are two important factors in the clinical KBQA task. 
\vspace{-2mm}


%% file: 6-appendix.tex
\section{Appendix}
\subsection{Comparisons of QA on Different Types of EMR Data}

Table~\ref{tab:question-example} in the main paper provides a comparison of questions that can be answered on different types of EMR data including clinical notes, structured tables and knowledge base. We can observe that knowledge base of patient clinical information is able to answer the basic questions that are answerable by QA on both clinical notes and structured tables. In addition, it has the ability to combine the advantages of free-text clinical notes and structured tables to handle more complex questions. In addition,  
Table~\ref{tab:data_comparisons} {shows a comparison of the ClinicalKBQA to the existing datasets for QA in healthcare.}

\begin{table}[htp]
\centering
\caption{Comparison of ClinicalKBQA with other datasets for QA in healthcare domain.}
\resizebox{\linewidth}{!}{
\begin{tabular}{l|ccc}\hline 
\bf Dataset & \bf Data Source & \bf QA Task &\bf Answer Type\\ \hline
BioASQ &Biomedical Articles &MRC & Text Span \\
CliCR & Clinical Reports &MRC &Text Span  \\
PubMedQA &Pubmed Abstracts &MRC & Text Span\\
emrQA &Clinical Notes &MRC & Text Span\\
MIMICSQL &Structured Tables & Text-to-SQL & Table Content\\\hline
ClinicalKBQA &Clinical Notes & KBQA &KB Entity\\
\hline
\end{tabular}
}
\label{tab:data_comparisons}
\end{table}

\begin{figure}[htp]
	\centering
	\includegraphics[width=\linewidth]{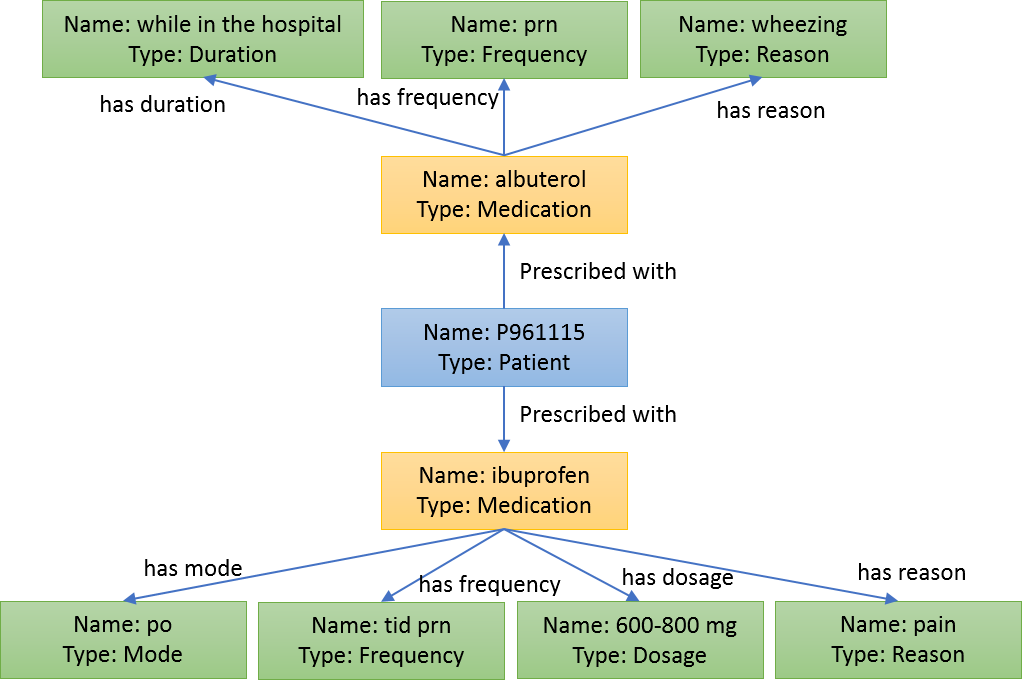}
	\caption{An example subgraph about prescribed medications along with their related information in \textit{medication} dataset.}
	\label{fig:subgraph-2}
\end{figure}

\begin{figure}[!tp]
	\centering
	\includegraphics[width=\linewidth]{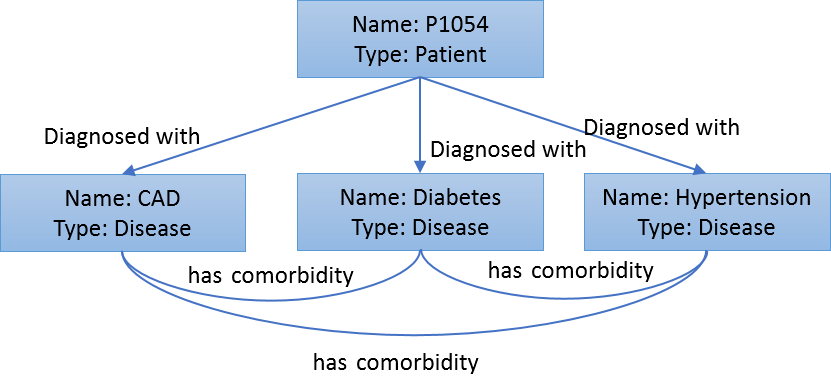}
	\caption{An example subgraph about diagnosed diseases and their comorbidity relationships in the Obesity dataset. A question based on this subgraph will be ``\textit{What is the comorbidities of diabetes for patient P1054}?''. The corresponding answers will be \textit{CAD} and \textit{Hypertension}.}
	\label{fig:subgraph-1}
\end{figure}

\subsection{More Details about ClinicalKBQA}
Here are the brief introduction about the annotations for various NLP task in n2c2.

\begin{itemize}[leftmargin=*]
\item \textbf{Smoking status classification} \cite{uzuner2008identifying}: Each clinical record is annotated with the smoking status from five possible categories (current smoker, past smoker, non-smoker, smoker, and unknown) along with the smoking-related facts mentioned in the records. 

\item \textbf{Identification of obesity and its co-morbidities} \cite{uzuner2009recognizing}: Each clinical record is annotated with obesity and co-morbidities using both textual judgments (explicit) and intuitive judgments (implicit). 

\item \textbf{Medication extraction} \cite{uzuner2010extracting}: The medication-related information including medication name, dosage along with the mode, frequency, duration and reason of the administration, is annotated in each clinical record.

\item \textbf{Analysis of relations of medical problems, tests and treatments} \cite{uzuner2010community}: The annotations for concept, assertion, and relation information are provided in each clinical record.

\item \textbf{Co-reference resolution} \cite{uzuner2012evaluating}: Each clinical record is annotated with concept mentions that are referring to the same entity. 

\item \textbf{Temporal information extraction and reasoning} \cite{sun2013evaluating}: The clinically significant events and temporal expressions are annotated along with the temporal relation between them in each clinical record. 

\item \textbf{Risk factors identification of heart disease}~\cite{stubbs2015identifying}: Each clinical record provides the annotation of medically relevant information about heart disease risk factors including the status of smoking, obesity, medication, and hypertension. 
\end{itemize}

\subsection{Subgraph Examples about ClinicalKB}


We provide a subgraph example in Obesity dataset about diagnosed diseases for patient P1054 and their comorbidity relationships in Figure \ref{fig:subgraph-1}. Based on the clinical note of patient P1054, he/she has been diagnosed with three diseases, including CAD, Diabetes and Hypertension. Since the annotations in Obesity dataset focus on the comorbidities relations of different diseases, we include such comorbidity relation between these three diseases.

\begin{figure*}[!tp]
\subfigure[]{\includegraphics[width=0.45\textwidth]{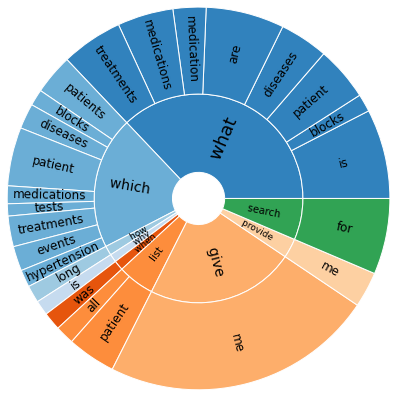}\label{fig:question-types-1}}
\hspace{4mm}
\subfigure[]{\includegraphics[width=0.45\textwidth]{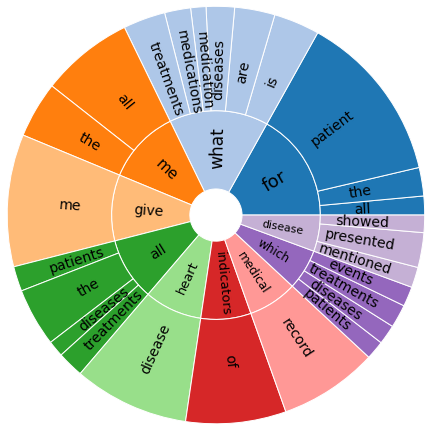}\label{fig:question-types-2}}
\caption{Distribution of questions by (a) the first two words in all questions and (b) the most common bigrams used in all questions.}
\label{fig:question-types}
\end{figure*}

\begin{table*}[!tp]
\centering
\caption{Question types in ClinicalKBQA along with examples.}
\resizebox{0.95\linewidth}{!}{
\begin{tabular}{c|l|c}\hline 
\bf Question type & \bf Examples & \bf Percentage \\ \hline
&What medications has patient P939003 ever been prescribed? & \\
What &What is the smoking status of patient P164? & 38.29\%  \\
&What is the dosage of colacefor patient P11995? &  \\\hline

&{List all comorbilities of Asthma for patient P1225.} &  \\
{List/Search/Give/Provide} &{Search for all the coreferenced tests of blood cultureon P727.} &35.21\%   \\
&{Give me all patients whose smoking status is current smoker.} &\\
&{Provide me the discharge time of patient P76.} &  \\\hline

&{Which tests are conducted on patient P0161?} & \\
Which &{Which tests are conducted on patient P0161?} &20.81\%   \\
&{Which medications can be prescribed for preventing creatinine?} &  \\\hline

&{Why is patient P74976 prescribed glucotrol?} &\\
Why &{Why is patient P280639 on coumadin?} & 2.21\%  \\
&{Why was ibuprofen originally prescribed for patient P961115?} & \\\hline

&{How much aspirin does patient P920102 take per day? } &  \\
{How much/often/long} &{How often does patient P439766 take regular  insulin?} &2.13\%   \\
&{How long has patient P652612 been taking levofloxacin?} & \\\hline

When &{When was patient P130 admitted?} &1.34\% \\
&{When was patient P32 discharged?} & \\\hline
\end{tabular}
}
\label{tab:question_types}
\end{table*}

Figure \ref{fig:subgraph-2} shows the relationships between patient P961115 and the prescribed medications along with other detailed attribute information including dosage, frequency, duration, and reason. We observe that not all attribute information is available for each medication. For example, the duration is only mentioned for albuterol, while the mode and dosage are mentioned only for ibuprofen. We hope that these two subgraph examples can provide an overview for understanding about patient information covered in ClinicalKB. Detailed statistics about ClinicalKB are summarized in Table \ref{tab:data-stat}.

\subsection{Question Distribution in ClinicalKBQA Data}
We group the questions in ClinicalKBQA dataset into different types based on the starting words. 
The distributions of question types showed in Figure~\ref{fig:question-types-1} are generated based on the most common first two starting words in all questions.
In Figure~\ref{fig:question-types-2}, we also show a distribution of the most common bigrams used in all questions in ClinicalKBQA dataset. It provides an overview about the specific patient information that various questions aim to extract from clinical notes. 
In addition, Table~\ref{tab:question_types} provides the quantitative percentage of various question types in ClinicalKBQA data.

\subsection{Model Framework}
\begin{figure}[htp]
	\centering
	\includegraphics[width=\linewidth]{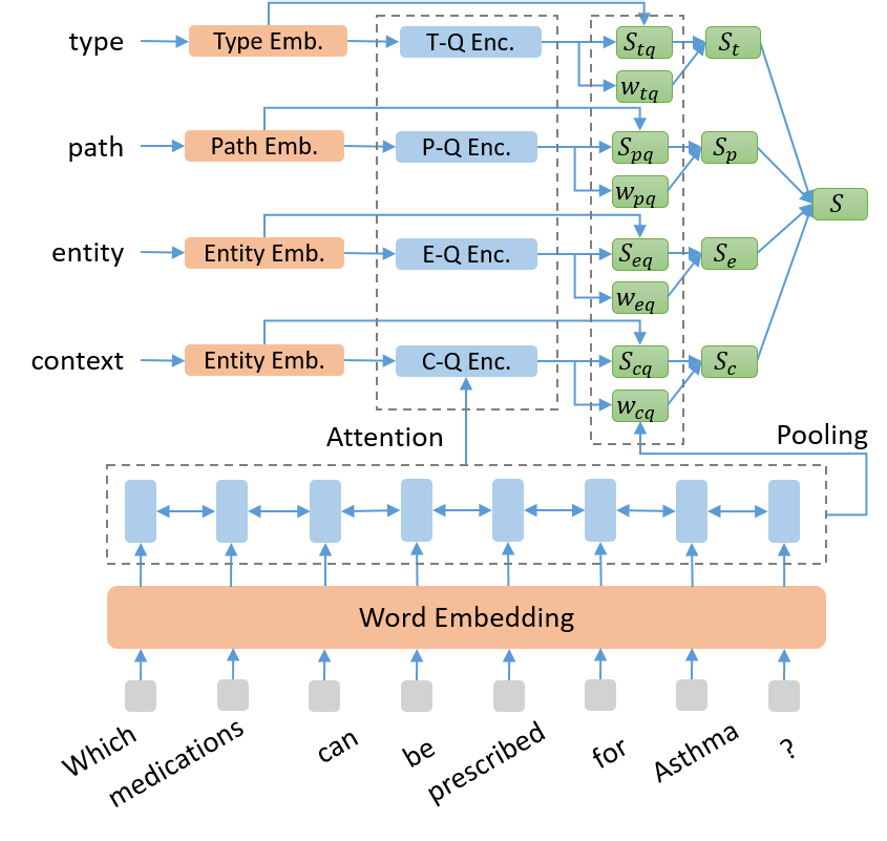}
	\caption{The overall framework of AAR model.}
	\label{fig:kbqa_model}
\end{figure}

Figure \ref{fig:kbqa_model} shows the overall framework of the AAR model.

